\documentclass[10pt,twocolumn,letterpaper]{article}

\usepackage{cvpr}
\usepackage{times}
\usepackage{epsfig}
\usepackage{graphicx}
\usepackage{amsmath}
\usepackage{amssymb}
\usepackage{mmstyle}
\usepackage[pagebackref=true,breaklinks=true,letterpaper=true,colorlinks,bookmarks=false]{hyperref}

\usepackage{nicefrac}       
\usepackage{microtype}      

\usepackage{multirow}
\usepackage{verbatim}
\usepackage{color}
\usepackage{float}
\usepackage{enumitem}
\usepackage{booktabs}
\usepackage{tabulary,multirow,overpic,xcolor}
\usepackage[caption=false]{subfig}

\usepackage[british,UKenglish,USenglish,english,american]{babel}
\newcolumntype{x}[1]{>{\centering\arraybackslash}p{#1pt}}

\newcommand{\app}{\raise.17ex\hbox{$\scriptstyle\sim$}}

\def\x{$\times$}

\newlength\savewidth\newcommand\shline{\noalign{\global\savewidth\arrayrulewidth
  \global\arrayrulewidth 1pt}\hline\noalign{\global\arrayrulewidth\savewidth}}
\newcommand{\tablestyle}[2]{\setlength{\tabcolsep}{#1}\renewcommand{\arraystretch}{#2}\centering\footnotesize}

\makeatletter\renewcommand\paragraph{\@startsection{paragraph}{4}{\z@}
  {.5em \@plus1ex \@minus.2ex}{-.5em}{\normalfont\normalsize\bfseries}}\makeatother

\def\cvprPaperID{4504} 

\cvprfinalcopy 
\begin{document}

\title{Temporal Pyramid Network for Action Recognition}

\author{
  Ceyuan Yang$^{\dag,1}$, Yinghao Xu$^{\dag,1}$, Jianping Shi$^2$, Bo Dai$^1$, Bolei Zhou$^1$\\
  $^1$The Chinese University of Hong Kong, $^2$SenseTime Group Limited\\
  {\tt\small \{yc019, xy119, bdai, bzhou\}@ie.cuhk.edu.hk},
  {\tt\small shijianping@sensetime.com}\\
}

\maketitle

\begin{abstract}
   Visual tempo characterizes the dynamics and the temporal scale of an action. Modeling such visual tempos of different actions facilitates their recognition.
Previous works often capture the visual tempo through sampling raw videos at multiple rates and constructing an input-level frame pyramid, which usually requires a costly multi-branch network to handle. 
In this work we propose a generic Temporal Pyramid Network (TPN) at the feature-level, 
which can be flexibly integrated into 2D or 3D backbone networks in a plug-and-play manner. Two essential components of TPN, the source of features and the fusion of features, form a feature hierarchy for the backbone so that it can capture action instances at various tempos. TPN also shows consistent improvements over other challenging baselines on several action recognition datasets.
Specifically, when equipped with TPN, the 3D ResNet-50 with dense sampling obtains a
2\% gain on the validation set of Kinetics-400.
A further analysis also reveals that TPN gains most of its improvements on action classes that have large variances in their visual tempos, validating the effectiveness of TPN.\footnote{Code and models are available at \href{https://decisionforce.github.io/TPN/}{this link.} \\ \indent $^\dag$ indicates equal contribution.}

\end{abstract}

\section{Introduction}
\label{sec:intro}
\begin{figure}[t]
    \centering
    \includegraphics[width=1.0\linewidth]{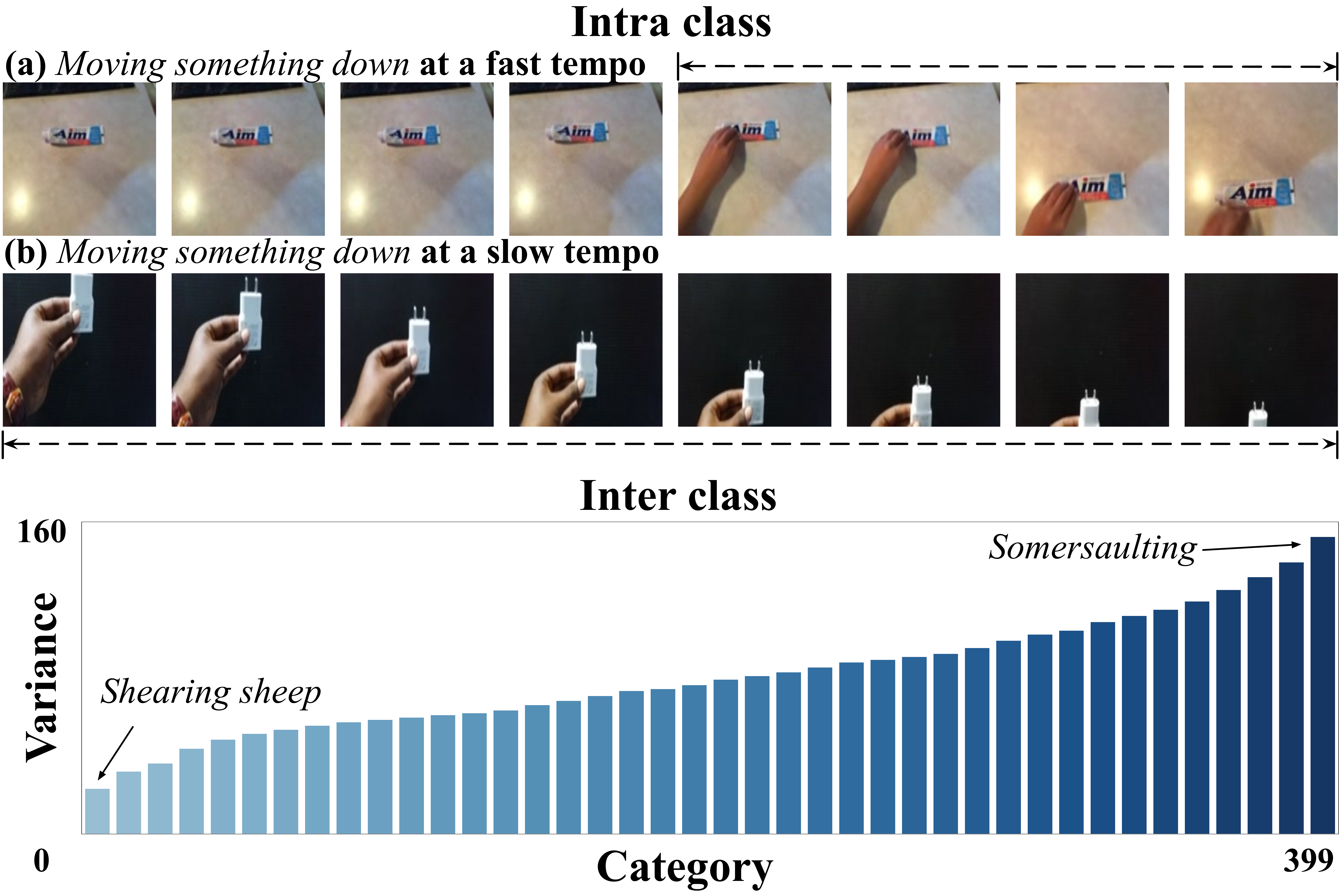}
    \caption{ 
        \textbf{Visual tempo variation of intra- and inter-class.}
        The action examples above show that people tend to act at different tempos even for the same action.
        The plot below shows different action categories sorted by their variances of visual tempos. 
        Specifically \emph{Somersaulting} has the largest variance in the visual tempo of its instances while \emph{Shearing sheep} has the smallest variance.
        Details of variation measurements can be found in the experiment section.
    }
    \label{fig:variation}
\end{figure}

While great progress has been made by deep neural networks to improve the accuracy of video action recognition \cite{slowfast, nonlocal, spn, trajconv, trajpool},
an important aspect of characterizing dfferent actions is often missed in the design of these recognition networks - the visual tempos of action instances. 
Visual tempo actually describes how fast an action goes, which tends to determine the effective duration at the temporal scale for recognition.
As shown at the bottom of Figure \ref{fig:variation}, action classes naturally have different visual tempos (\eg~\emph{hand clapping} and \emph{walking}).
In some cases the key to distinguish different action classes is their visual tempos, as they might share high similarities in visual appearance, \
such as \emph{walking}, \emph{jogging} and \emph{running}.
Moreover, as shown at the top of Figure \ref{fig:variation}, when performing the same action, each performer may act at his/her own visual tempo, due to various factors such as age, mood, and energy level.
\eg~,an elder man tends to move slower than a younger man, so as a man with a heavier weight.
Precise modeling of such intra- and inter-class variances in visual tempos of action instances can potentially bring a significant improvement to action recognition.

Previous attempts \cite{slowfast, dtpn, spn} for extracting the dynamic visual tempos of action instances mainly rely on constructing a frame pyramid, 
where each pyramid level samples the input frames at a different temporal rate.
For example, we can sample from the total $64$ frames of an video instance at intervals $16$ and $2$ respectively, to construct a two-level frame pyramid consisting of $4$ and $32$ frames.
Subsequently, frames at each level are fed into different backbone subnetworks, and their output features are further combined together to make the final prediction. 
By sampling frames at different rates as input, backbone networks in \cite{slowfast, dtpn} are able to extract features of different receptive fields and represent the input action instance at different visual tempos.
These backbone subnetworks thus jointly aggregate temporal information of both fast-tempo and slow-tempo, handling action instances at different temporal scales.

Previous methods \cite{slowfast, dtpn, spn} have obtained noticeable improvements for action recognition, however it remains computationally expensive to deal with the dynamic visual tempos of action instances at the input frame level. 
It is not scalable to pre-define the tempos in the input frame pyramid and then feed the frames into multiple network branches, especially when we use a large number of sampling rates.
On the other hand, many commonly-used models in video recognition, such as C3D and I3D \cite{c3d,kinetics}, often stack a series of temporal convolutions. In these networks, as the depth of a layer increases, its temporal receptive field increases as well.
As a result, the features at different depths in a \emph{single} model already capture information of both fast-tempo and slow-tempo.
Therefore, we propose to build a temporal pyramid network (TPN) to aggregate the information of various visual tempos at \textit{feature level}.
By leveraging the feature hierarchy formed inside the network, the proposed TPN is able to work with input frames fed at a single rate. 
As an auxiliary module, TPN could be applied in a plug-and-play manner to various existing action recognition models to bring consistent improvements.

In this work we first provide a general formulation of the proposed TPN, where several components are introduced to better capture the information at multiple visual tempos.
We then evaluate TPNs on three benchmarks: Kinetics-400 \cite{kinetics}, Something-Something V1 \& V2 \cite{sthv1} and Epic-Kitchen \cite{epic} with comprehensive ablation studies.
Without any bells and whistles, TPNs bring consistent improvements when combined with both 2D and 3D networks.
Besides, the ablation study shows that TPN obtains most of its improvements from the action classes that have significant variances in visual tempos.
This result verifies our assumption that aggregating features in a single model is sufficient to capture the visual tempos of action instances for video recognition.

\section{Related Work}
\label{sec:relat}

\paragraph{Video Action Recognition.}

Attempts for video action recognition could be divided into two categories.
Methods in the first category often adopt a 2D + 1D paradigm,\
where 2D CNNs are applied over per-frame inputs, followed by a 1D module that aggregates per-frame features.
Specifcally, two-stream networks in \cite{twostream,tsfusion,cts,cnnforar} utilize two separate CNNs on per-frame visual appearances and optical flows respectively,
and an average pooling operation for temporal aggregation.
Among its variants, TSN \cite{tsn} proposes to represent video clips by sampling from evenly divided segments.
And TRN \cite{trn} and TSM \cite{tsm} respectively replace the average pooling operation with an interpretable relational module and utilize a shift module,\
in order to better capture information along the temporal dimension.
However, due to the deployment of 2D CNNs in these methods, semantics of the input frames could not interact with each other in the early stage,\
which limits their ability to capture the dynamics of visual tempos.
Methods \cite{c3d,3dforar} in the second category alternatively apply 3D CNNs that stack 3D convolutions to jointly model temporal and spatial semantics.
Along this line of research, Non-local Network \cite{nonlocal} introduces a special non-local operation \
to better exploit the long-range temporal dependencies between video frames.
Besides Non-local Network, different modifications for the 3D CNNs,\
including the inflating 2D convolution kernels \cite{kinetics} and the decomposing 3D convolution kernels \cite{p3d, r21d, r21d_v2},\
can also boost the performances of 3D CNNs.
Other effects \cite{trajpool,trajconv,improvedtraj,shao2020finegym,shao2020tapos} are taken on irregular convolution/pool for better feature alignment or study action instances in a fine-grained way. 
Although the aforementioned methods could better handle temporal information,\
the large variation of visual tempos remains neglected.

\paragraph{Visual Tempo Modeling in Action Recognition.}
The complex temporal structure of action instances, particularly in terms of the various visual tempos, raises a challenge for action recognition.
In recent years, researchers have started exploring this direction.
SlowFast \cite{slowfast} hard-codes the variance of visual tempos using an input-level frame pyramid that has level-wise frames sampled at different rates.
Each level of the pyramid is also separately processed by a network, where mid-level features of these networks are interactively combined.
With the assist of both the frame pyramid and the level-specific networks, SlowFast could robustly handle the variance of visual tempos.
The complex temporal structure inside videos, particularly tempo variation, raises a challenge for action recognition.
DTPN \cite{dtpn} also samples frames with different frame per seconds (FPS) to construct a natural pyramidal representation for arbitrary-length input videos.
However, such a hard-coding scheme tends to require multiple frames , especially when the pyramid scales up.
Different from previous feature-level pyramid networks \cite{Hypercolumns, fpn, panet, li2018feature} which deal with the large variance of spatial scales in object detection, we instead leverage the feature hierarchy to handle the variance of temporal information \ie visual tempos. 
In this way we could hide the concern about visual tempos inside a single network,\
and we only need frames sampled at a single rate at the input-level.

\begin{figure*}[t]
    \centering
    \begin{tabular}{@{\hspace{0mm}}c}
    \includegraphics[width=1.0\linewidth]{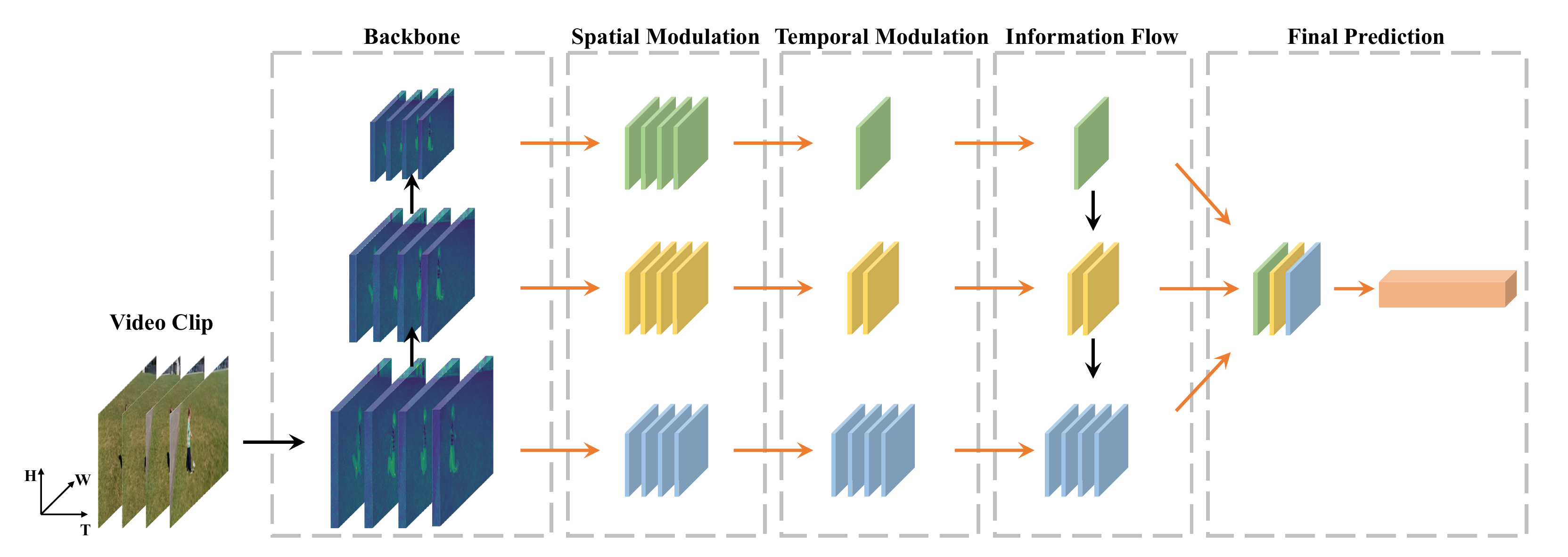}
    \end{tabular}
    \caption{
        \textbf{Framework of TPN:}
        \emph{Backbone Network} to extract multiple level features.
        \emph{Spatial Semantic Modulation} spatially downsamples features to align semantics.
        \emph{Temporal Rate Modulation} temporally downsamples features to adjust relative tempo among levels.
        \emph{Information Flow} aggregates features in various directions to enhance and enrich level-wise representations.
        \emph{Final Prediction} rescales and concatenates all levels of pyramid along channel dimension. 
        Note that the channel dimensions in \emph{Final Prediction} and corresponding operations are omitted for brevity. 
    }
    \label{fig:framework}
    \end{figure*}

\section{Temporal Pyramid Network}
\label{sec:method}

The visual tempo of an action instance is one of the key factors for recognizing it, especially when other factors are ambiguous. 
For example, we cannot tell if an action instance belongs to \emph{walking, jogging} or \emph{running} based on its visual appearance.
However, 
it is difficult to capture the visual tempos due to their inter- and intra-class variance across different videos. 
Previous works \cite{slowfast,dtpn,spn} address this issue at the input-level.
They utilize a frame pyramid that contains frames sampled at pre-defined rates to represent the input video instance at various visual tempos.
Since each level of the frame pyramid requires a separate backbone network to handle,\
such an approach may be computationally expensive, especially when the level of pyramid scales up.

Inspired by the observation that features at multiple depths in a \emph{single} network already cover various visual tempos,\
we propose a feature-level temporal pyramid network (TPN) for modeling the visual tempo.
TPN could operate on only a single network no matter how many levels are included in it.
Moreover, TPN could be applied to different architectures in a plug-and-play manner.
To fully implement TPN, two essential components of TPN must be designed properly,\
namely 1) the feature source and 2) the feature aggregation.
We propose the spatial semantic modulation and temporal tempo modulation to control the relative differences of the feature source in Sec.\ref{subsec:sources}, 
and construct multiple types of information flows for feature aggregation in Sec.\ref{subsec:infoflow}.
Finally we show how to adopt TPN for action recognition in Sec.\ref{subsec:application},
taking \cite{slowfast} as an exemplar backbone network.

\subsection{Feature Source of TPN}\label{subsec:sources}

\paragraph{Collection of Hierarchical Features.}
While TPN is built upon a set of $M$ hierarchical features that have increasing temporal receptive fields from bottom to top,\
there are two alternative ways to collect these features from a backbone network.
1) \emph{Single-depth pyramid}: a simple way is to choose a feature $\mF_\mathrm{base}$ of size $C \times T \times W \times H$ at some depth,\
and to sample along the temporal dimension with $M$ different rates $\{r_1, ..., r_M;r_1 < r_2 < ... < r_M\}$.
We refer to such a TPN as a single-depth pyramid consisting of $\{\mF_\mathrm{base}^{(1)}, ..., \mF_\mathrm{base}^{(M)}\}$ of sizes $\{C \times \frac{T}{r_1} \times W \times H, ..., C \times \frac{T}{r_M} \times W \times H\}$.
Features collected in this way could lighten the workload of fusion as they have identical shapes besides the temporal dimension.
However, they may limit in effectiveness as they represent video semantics only at a single spatial granularity.
2) \emph{Multi-depth pyramid}: a better way is to collect a set of $M$ features with increasing depths,\
resulting in a TPN made of $\{\mF_1, \mF_2, ..., \mF_M\}$ of sizes $\{C_1 \times T_1 \times W_1 \times H_1, ..., C_M \times T_M \times W_M \times H_M\}$,
where generally the dimensions satisfy $\{C_{i_1} \ge C_{i_2}, W_{i_1} \ge W_{i_2}, H_{i_1} \ge H_{i_2}; i_1 < i_2\}$. 
Such a multi-depth pyramid contains richer semantics in the spatial dimensions,\
yet raises the need of careful treatment in feature fusion,
in order to ensure correct information flows between features.

\paragraph{Spatial Semantic Modulation.} 

To align spatial semantics of features in the multi-depth pyramid, a spatial semantic modulation is utilized for TPN.\
The spatial semantic modulation works in two complementary ways.
For each but the top-level feature, a stack of convolutions with level-specific stride are applied to it,\
matching its spatial shape and receptive field with the top one.
Moreover, an auxiliary classification head is also appended to it to receive stronger supervision, leading to enhanced semantics.
The overall objective for a backbone network with our proposed TPN thus becomes:
\begin{align}
	\cL_\mathrm{total} = \cL_{CE,o} + \sum_{i=1}^{M-1} \lambda_i \cL_{CE,i},
\end{align}
where $\cL_{CE,o}$ is the original Cross-Entropy loss, and $\cL_{CE, i}$ is the loss for $i$-th auxiliary head. $\{\lambda_i\}$ are balancing coefficients. 
After spatial semantic modulation, features have aligned shapes and consistent semantics in the spatial dimensions.
However, it remains uncalibrated in the temporal dimension, where we introduce the proposed temporal rate modulation.

\paragraph{Temporal Rate Modulation.} 

Recall in the input-level frame pyramid used in \cite{slowfast}, the sampling rates of frames could be adjusted dynamically to increase its applicability.
On the contrary, TPN is limited in the flexibility, as it operates on features of a backbone network,\
so that the visual tempos of these features are only controlled by their depths in the original network.
To equip TPN with a similar flexibility as in the input-level frame pyramid,\
a set of hyper-parameters $\left\{\alpha_i\right\}_{i=1}^M$ are further introduced to TPN for temporal tempo modulation.
Specifically, $\alpha_i$ denotes that after spatial semantic modulation,\
the updated feature at $i$-level will be temporally downsampled by a factor of $\alpha_i$, using a parametric sub-net.
The inclusion of such hyper-parameters enables us to better control the relative differences of features in terms of temporal scales,\
so that feature aggregation could be conducted more effectively.
With some abuse of notations, we refer to $\mF_i$ of size $C_i \times T_i \times W_i \times H_i$ as the $i$-th feature after both the spatial semantic modulation and the temporal rate modulation in the following content.

\subsection{Information Flow of TPN}\label{subsec:infoflow}

\begin{figure}[t]
    \centering
    \includegraphics[width=1.0\linewidth]{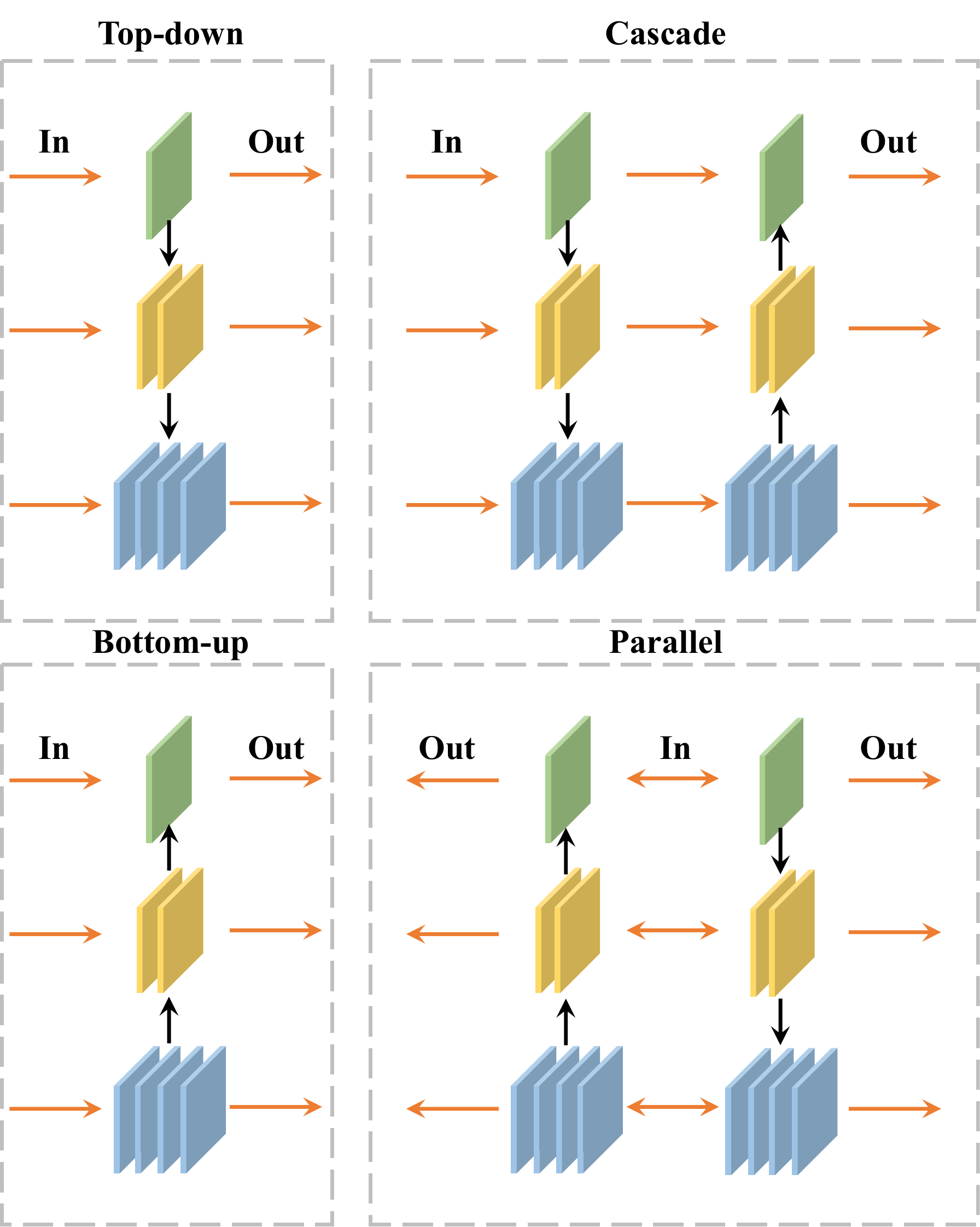}
    \caption{
        \textbf{Information Flow:}
        \textbf{Black arrows} illustrate the aggregation directions while the \textcolor{orange}{orange arrows} denote the IO stream from \emph{Temporal Modulation} to \emph{Final Prediction} of Figure \ref{fig:framework}.
        The channel dimensions and up/downsample operations are omitted.
    }
    \label{fig:flow}
    \end{figure}

After collecting and pre-processing the hierarchical features as in Sec.\ref{subsec:sources},\
so that they are dynamic in visual tempos and consistent in spatial semantics,\
we are ready to step in the second step of TPN construction -- how to aggregate these features.
Let $\mF'_i$ be the aggregated feature at $i$-th level, generally there are three basic options:
\begin{align}
	\mF'_i = \begin{cases}
			\mF_i & \mathrm{Isolation~Flow} \\
			\mF_i \bigoplus g(\mF_{i-1}, T_i/T_{i-1}) & \mathrm{Bottom}$-$\mathrm{up~Flow} \\
			\mF_i \bigoplus g(\mF_{i+1}, T_i/T_{i+1}) & \mathrm{Top}$-$\mathrm{down~Flow}
		\end{cases},
\end{align} 
where $\bigoplus$ denotes element-wise addition.
And to ensure the compatibility of the addition between consecutive features,
during aggregation a down/up-sampling operation, $g(\mF, \delta)$ with $\mF$ as the feature and $\delta$ is the factor, is applied along the temporal dimension.
Note that the top/bottom features for top-down/bottom-up flow would not be aggregated by other features.
Besides the above basic flows to aggregate features in TPN, we could also combine them to achieve two additional options, namely \emph{Cascade Flow} and \emph{Parallel Flow}.
While applying a bottom-up flow after a top-down flow will lead to the cascade flow, applying them simultaneously will result in the parallel flow.
See Fig. \ref{fig:flow} for an illustration of all the possible flows. 
It's worth noting that more complicated flow (\eg~path aggregation in \cite{panet}) could be built on top of these flows.
However, our attempts in this line of research have not shown further improvement.
Finally, following Fig.\ref{fig:framework}, all aggregated features in TPN will be rescaled and concatenated for succeeding predictions.

\subsection{Implementation}\label{subsec:application}
Here we introduce the implementation of TPN for action recognition.
Following \cite{slowfast}, we use inflated ResNet \cite{slowfast} as the 3D backbone network, for its promising performance on various datasets \cite{kinetics}.
Meanwhile, original ResNet \cite{resnet} serves as our 2D backbone.
We use the output features of \emph{res2, res3, res4, res5} to build TPN,
where they are spatially downsampled by respectively $4, 8, 16$ and $32$ times, compared to the input frames.
We provide the structure of 3D ResNet-50 in Tab.\ref{table:arch} for the reference.
In the spatial semantic modulation, a stack of convolutions with $M-i$ stride to process the feature at $i$-th level in a $M$-level TPN and the feature dimension would be decreased or increased to 1024.
Besides, the temporal rate modulation for each feature is achieved by a convolutional layer and a max-pooling layer.
Finally, after feature aggregation through one of the five flows mentioned in Sec. \ref{subsec:infoflow},\
features of TPN will be separately rescaled by max-pooling operations,\
and their concatenation will be fed into a fully-connected layer to make the final predictions.
TPN can be also jointly trained with the backbone network in an end-to-end manner.

\section{Experiments}
\label{sec:exper}
\newcommand{\blockb}[3]{\multirow{3}{*}{\(\left[\begin{array}{c}\text{1\x1\x1, #2}\\[-.1em] \text{1\x3\x3, #2}\\[-.1em] \text{1\x1\x1, #1}\end{array}\right]\)\x#3}}
\newcommand{\blockc}[3]{\multirow{3}{*}{\(\left[\begin{array}{c}\text{3\x1\x1, #2}\\[-.1em] \text{1\x3\x3, #2}\\[-.1em] \text{1\x1\x1, #1}\end{array}\right]\)\x#3}}
    \begin{table}[t]
    \footnotesize
    \centering
    \resizebox{1.0\columnwidth}{!}{
    \tablestyle{6pt}{1.08}
    \begin{tabular}{c|c|c}
    
    Stage & Layer & Output size \\
    \shline
    raw   & - & 8 \x 224 \x 224 \\ \hline
    conv$_1$ & \multicolumn{1}{c|}{1\x7\x7, 64, stride 1, 2, 2} & 8\x112\x112 \\
    \hline
    pool$_1$  & \multicolumn{1}{c|}{1\x3\x3 max, stride 1, 2, 2} & 8\x56\x56 \\
    \hline
    \multirow{3}{*}{res$_2$} & \blockb{256}{64}{3} & \multirow{3}{*}{8\x56\x56} \\
      &  & \\
      &  & \\
    \hline
    \multirow{3}{*}{res$_3$} & \blockb{512}{128}{4} & \multirow{3}{*}{8\x28\x28} \\
      &  & \\
      &  & \\
    \hline
    \multirow{3}{*}{res$_4$} & \blockc{1024}{256}{6} & \multirow{3}{*}{8\x14\x14}  \\
      &  & \\
      &  & \\
    \hline
    \multirow{3}{*}{res$_5$} & \blockc{2048}{512}{3} & \multirow{3}{*}{8\x7\x7} \\
      &  & \\
      &  & \\
    \hline
    \multicolumn{2}{c|}{global average pool, fc} & 1\x1\x1  \\
    \end{tabular}}
    \vspace{.5em}
    \caption{
    \textbf{3D Backbone.}    
    Following \cite{slowfast}, our inflated 3D ResNet-50 backbone for video is shown. Note that both output size and kernel size are in T\x W\x H shape.
    }
    \vspace{-1em}
    \label{table:arch}
    \end{table}

We evaluate the proposed TPN on various action recognition datasets, \
including Kinetics-400 \cite{kinetics}, Something-Something V1 \& V2 \cite{sthv1}, and Epic-Kitchen \cite{epic}. The consistent improvements show the effectiveness and generality of TPN.
Ablation studies on the components of TPN are also included.
Moreover, we present several empirical analysis to verify our motivation of TPN,\
\ie~a feature-level temporal pyramid on a single backbone is beneficial for capturing the variance of visual tempos.
All experiments are conducted with the single modality (\ie RGB frames) on MMAction \cite{mmaction2019} and evaluated on the validation set unless specified.

\paragraph{Dataset.}
Kinetics-400 \cite{kinetics} contains around 240k training videos and 19k validation videos that last for 10 seconds. It includes 400 action categories in total.
Something-Something V1 \cite{sthv1} consists of 86k training videos and 11k validation videos belonging to 174 action categories, whose durations vary from 2 to 6 seconds.
The second release (V2) of Something-Something increase the number of videos to 220k.
Epic-Kitchen \cite{epic} includes around 125 verb and 352 noun categories. Following \cite{anticipation}, we randomly select 232 videos (23439 segments) for training and 40 videos (4979 segments) for validation.

\paragraph{Training.}
Unless specified otherwise, our models are defaultly initialized by pre-trained models on ImageNet \cite{imagenet}.
Following the setting in \cite{slowfast}, the input frames are sampled from a set of consecutive 64 frames at a specific interval $\tau$.
Each frame is randomly cropped so that its short side ranges in $\left[256, 320\right]$ pixels, as in \cite{nonlocal,slowfast,vggnet}.
The augmentation of horizontal flip and a dropout \cite{dropout} of 0.5 are adopted to reduce overfitting.
And BatchNorm (BN) \cite{bn} is not frozen.
We use a momentum of 0.9, a weight decay of 0.0001 and a synchronized SGD training over 8 GPUs \cite{sgd1hour}.
Each GPU has a batch-size of 8, resulting in a mini-batch of 64 in total.
For Kinetics-400, the learning rate is 0.01 and will be reduced by a factor of 10 at 100, 125 epochs (150 epochs in total) respectively.
For Something-Something V1 \& V2 \cite{sthv1} and Epic-Kitchen \cite{epic}, our model is trained for 150 and 55 epochs separately. 

\paragraph{Inference.}
There exist two ways for inference: \emph{three-crop} and \emph{ten-crop} testing.
a) \emph{Three-crop} testing refers to three random crops of size $256\times256$ from the original frames, which are resized firstly to have 256 pixels in their shorter sides.
\emph{Three-crop} testing is used as the approximation of spatially fully-convolutional testing as in \cite{vggnet, nonlocal, slowfast}. 
b) \emph{Ten-crop} testing basically follows the procedure of \cite{tsn}, which extracts 5 crops of size $224\times224$ and flips these crops.
Specially, we conduct \emph{three-crop} testing on Kinetics-400. We also uniformly sample 10 clips of the whole video and average the softmax probabilities of all clips as the final prediction.
For the other two datasets, \emph{ten-crop} testing and TSN-like methods with 8 segments are adopted. 

\paragraph{Backbone.}
We evaluate TPN on both 2D and 3D backbone networks.
Specifically, the \emph{slow-only} branch of SlowFast \cite{slowfast} is applied as our backbone network (denoted as I3D) due to its promising performance on various datasets. 
The architecture of I3D is shown in Table \ref{table:arch}, which turns the 2D ResNet \cite{resnet} into a 3D version via inflating kernels \cite{nonlocal, kinetics}.
Specifically, a 2D kernel of size $k \times k$ will be inflated to have the size $t \times k \times k$, with its original weights copied for $t$ times and rescaled by $1/t$.
Note that there are no temporal downsampling operations in the \emph{slow-only} backbone.
ResNet-50 \cite{resnet} is used as 2D backbone to show that TPN could combine with various backbones.
The final prediction follows the standard protocol of TSN \cite{tsn} unless specified.
%

\begin{table}[t]
    \begin{center}
        \setlength{\tabcolsep}{2.0mm}{
            \begin{tabular}{lcccc}
                \hline 
                Model                           &Frames & Flow  & Top-1 & Top-5       \\ \hline
                R(2+1)D \cite{r21d}             & 16    & \checkmark  & 73.9  & 90.9    \\
                I3D \cite{kinetics}             & 16    & \checkmark  & 71.6  & 90.0                \\
                Two-Stream I3D \cite{kinetics}  & 64    & \checkmark  & 75.7  & 92.0             \\ 
                S3D-G \cite{r21d_v2}            & 64    & \checkmark  & 77.2  & 93.0    \\  \hline
                STC-X101 \cite{stc}             & 32    &       & 68.7  & 88.5   \\
                Nonlocal-R50 \cite{nonlocal}    & 32    &       & 76.5  & 92.6                \\       
                Nonlocal-R101 \cite{nonlocal}   & 32    &       & 77.7  & 93.3                \\         
                SlowFast-R50 \cite{slowfast}    & 32    &       & 77.0  & 92.6    \\
                SlowFast-R101 \cite{slowfast}   & 32    &       & 77.9  & 93.2    \\ 
                CSN-101  \cite{cscn}            & 32    &       & 76.7  & 92.3    \\
                CSN-152  \cite{cscn}            & 32    &       & 77.8  & 92.8    \\
                \hline 
                \textbf{TPN-R50}                         & $32 \times 2$    &       & 77.7  & 93.3     \\
                \textbf{TPN-R101}                        & $32 \times 2$    &       & \textbf{78.9}  & \textbf{93.9}     \\
                \hline 
            \end{tabular}}
    \end{center}
    \caption{
    \textbf{Comparison with other state-of-the-art methods on the validation set of Kinetics-400.} Note that R50 and R101 denote the backbone networks and their depth respectively.}
    \label{table:results_kinetics}
\end{table}

\subsection{Results}\label{subsec:main_results}
\paragraph{Results on Kinetics-400.}

We compare our TPN with other state-of-the-art methods on Kinetics-400.
The \emph{multi-depth} pyramid and the parallel flow are used as the default setting for TPN. 
In detail, the multi-depth pyramid is built on the outputs of \emph{res4} and \emph{res5}.
And the hyper-parameters $\left\{\alpha_i\right\}_{i=1}^M$ are set to be $\left\{16,32\right\}$.
As discussed in the spatial semantic modulation, an additional auxiliary head is applied on the output of res4 with a balancing coefficient of $0.5$.
Sampling intervals of input frames $\tau=8,4,2$ are compared.

The performance of I3D-R50 + TPN (\ie~TPN-R50) is included in Table \ref{table:results_kinetics}.
It is worth noting that in Table \ref{table:results_kinetics} backbones of methods with the same depth are slightly different, which also affect their final accuracies.
TPN-R50 could achieve 77.7\% top-1 accuracy, better than others with the same depth.
TPN-R101 are also evaluated with the input setting of $32\times 2$,\
which obtains an accuracy of 78.9\%, surpassing other methods with the same numbers of input frames.

\begin{table}[t]
    \begin{center}
        \setlength{\tabcolsep}{2.8mm}{
            \begin{tabular}{lccc}
                \hline 
                Backbone                &  Segments  & Testing             & Top-1      \\ \hline
                TSN-50 from \cite{tsm}  &  8         & \emph{ten-crop}     & 69.9       \\
                TSM-50 from \cite{tsm}  &  8         & \emph{ten-crop}     & 72.8       \\ \hline
                TSN-50 + TPN            &  8         & \emph{ten-crop}     & \textbf{73.5}       \\ \hline
            \end{tabular}}
    \end{center}
    \caption{
        \textbf{Improvement of 2D backbone on the validation set of Kinetics-400.} Only 8 segments are used for both training and validation for apple-to-apple comparison.}
    \label{table:tsn_kinetics}
\end{table}

Being a general module, TPN could be combined with 2D networks.
To show this, we add TPN to the ResNet-50 \cite{resnet} in TSN (TSN-50 + TPN), and train such a combination with 8 segments (uniform sampling) in an end-to-end manner.
Different from the original TSN \cite{tsn} which takes 25 segments for validation, we utilize only 8 segments and the \emph{ten-crop} testing, comparing apples to apples.
As shown in Table \ref{table:tsn_kinetics}, adding TPN to TSN-50 could boost the top-1 accuracy by 3.6\%.

\paragraph{Results on Something-Something.}
\begin{table}[t]
    \begin{center}
        \setlength{\tabcolsep}{1mm}{
            \begin{tabular}{lcccc}
                \hline 
                   Backbone                    & Segments & Top1@V1        & Top1@V2       \\ \hline
                   TRN-Multiscale \cite{tsm}   & 8  & 38.9           & 48.8           \\ 
                   ECO \cite{eco}              & 8  & 39.6           & -              \\ 
                   TSN-50 \cite{tsm}           & 8  & 19.7           & 30.0                \\ 
                   TSM-50 \cite{tsm}           & 8  & 45.6           & 59.1                  \\ \hline
                   TSN-50 + TPN                & 8  & 40.6           & 55.2               \\ 
                   TSM-50 + TPN                & 8  & \textbf{49.0}  & \textbf{62.0}   \\ \hline
            \end{tabular}}
    \end{center}
    \caption{
        \textbf{Results on the validation set of Something-Something V1 \& V2.} Note that results on V1 \& V2 take the center crop of 1 clip/video according to \cite{tsm}.
        }
    \label{table:sth}
\end{table}
Results of different baselines with and without TPN on the Something-Something are also included in Table \ref{table:sth}.
For a fair comparison, we use the center crop of size $224\times224$ in all 8 segments, following the protocol used in TSM \cite{tsm}.
Both TSN and TSM receive a significant performance boost after combined with the proposed TPN.
While TSM has a relatively larger capacity compared to TSN, such consistent improvements on both backbones clearly demonstrate the generality of TPN.
Besides, on the \href{https://20bn.com/datasets/something-something/v2}{leaderboard} (dated on 04/10/2020), TPN with backbone of TSM-101$_{16f}$ achieves 67.7\% Top-1 accuracy, following the standard protocol \ie full resolution of 2 clips.

\paragraph{Results on Epic-Kitchen.}
\begin{table}[t]
    \begin{center}
        \setlength{\tabcolsep}{2.0mm}{
            \begin{tabular}{lccc}
                \hline 
                Model                        & Frames & NOUN@1 & VERB@1  \\ \hline
                TSN (RGB)      \cite{epic}   & 25     & 36.8   & 45.7     \\ 
                TSN (Flow)     \cite{epic}   & 25     & 17.4   & 42.8         \\ 
                TSN (Fusion)   \cite{epic}   & 25     & 36.7   & 48.2        \\ \hline
                TSN (our impl.)              & 8      & 39.7   & 48.2         \\
                TSN + TPN                    & 8      & \textbf{41.3}   & \textbf{61.1}        \\ \hline
            \end{tabular}}
    \end{center}
    \caption{
    \textbf{Results on the validation set of Epic-Kitchen.}    
    TSN is equipped with TPN.}
    \label{table:epic_ar}
\end{table}

As shown in Table \ref{table:epic_ar}, we compare TSN+TPN to two baselines on Epic-Kitchen, following the settings in \cite{epic}.
Consequently, a similar improvement is observed as in other datasets, especially on verb classification, which has an increase of $12.9\%$.

\begin{table*}[t]\centering\vspace{-1em}
    \captionsetup[subfloat]{captionskip=2pt}
    \captionsetup[subffloat]{justification=centering}
    \subfloat[
        \textbf{Possible Sources:} None denotes the I3D baseline with the depth of 50. \emph{res}$\left\{i\right\}$ means features are collected from the $i$-th stage in ResNet \cite{resnet}. Specially, \emph{res}$\left\{5\right\}$ takes the \emph{single-level} pyramid as \ref{subsec:sources}.
\label{table:ablations:sources}]{
    \setlength{\tabcolsep}{5mm}{
        \begin{tabular}{lcc}
            \hline
            Possible Sources                     & Top-1         & Top-5            \\ \hline
            None                                 & 74.9          & 91.9                           \\ 
            \emph{res}$\left\{2,3,4,5\right\}$   & 74.6          & 91.8                           \\
            \emph{res}$\left\{3,4,5\right\}$     & 74.9          & 92.1                           \\
            \emph{res}$\left\{4,5\right\}$       & \textbf{76.1} & \textbf{92.5}                  \\
            \emph{res}$\left\{5\right\}$         & 75.7          & 92.3                            \\ \hline
            \\
            \\
\end{tabular}}
    }\hspace{12mm}
    \subfloat[
        \textbf{Ablation Study on TPN components:}
        We gradually add auxiliary head (Head), spatial convolutions in semantic modulation (Spatial), temporal rate modulation (Temporal) and information flow (Flow) onto the baseline model.
        \label{table:ablations:component}
        ]{
            \setlength{\tabcolsep}{3.0mm}{
            \begin{tabular}{cccc|c}
                \hline
                Head         & Spatial      & Temporal   & Flow        & Top-1       \\ \hline
                             &              &            &             &  74.9       \\
                \checkmark   &              &            &             &  74.6       \\
                \checkmark   & \checkmark   &            &             &  75.2       \\ 
                \checkmark   & \checkmark   & \checkmark &             &  75.4       \\
                \checkmark   & \checkmark   & \checkmark & \checkmark  &  \textbf{76.1}       \\ \hline
                             & \checkmark   & \checkmark & \checkmark  &  75.9       \\ 
                             &              & \checkmark & \checkmark  &  75.6       \\  \hline
            \end{tabular}
            }
    }\hspace{16mm}
    \subfloat[  
        \textbf{Information Flow:} 
        Accuracy of several TPN variants mentioned in Sec \ref{subsec:infoflow} are shown. The hyper-parameters $\left\{\alpha_i\right\}_{i=1}^M$ is set as $\left\{4,8\right\}$.
        \label{table:ablations:flow}
        ]{
            \setlength{\tabcolsep}{5mm}{
            \begin{tabular}{lcc}
                \hline
                Information Flow          & Top-1           & Top-5               \\ \hline
                Isolation                 & 75.4            & 92.4                          \\
                Bottom-up                 & 75.8            & 92.3                           \\
                Top-down                  & 75.7            & 92.3                          \\
                Cascade                   & 75.9            & 92.3                          \\
                Parallel                  & \textbf{76.1}   & \textbf{92.5}                 \\ \hline
                \\
    \end{tabular}}
    }\hspace{10mm}
    \subfloat[
        \textbf{Input Frames:}
        Different number of frames are adopted to evaluate whether TPN has the consistent improvement. 
        \label{table:ablations:ninput}]{

        \begin{tabular}{lcccc}
            \hline
            Backbone     & $T \times \tau$               & w/o TPN     &   w/ TPN       & $\Delta$ Acc        \\ \hline
                         & 8$\times$8                & 74.9        &   76.1         &  +1.2                      \\
            I3D-50       & 16$\times$4               & 76.1        &   77.3         &  +1.2                    \\
                         & 32$\times$2               & 75.7        &   \textbf{77.7}&  +2.0                    \\ \hline
                         & 8$\times$8                & 76.0        &   77.2         &  +1.2                     \\
            I3D-101      & 16$\times$4               & 77.0        &   78.1         &  +1.1                     \\
                         & 32$\times$2               & 77.4        &   \textbf{78.9}&  +1.5                     \\ \hline
\end{tabular}
    }
    
    \vspace{1em}
    \caption{
        \textbf{Ablation studies on Kinetics-400.}
        Backbone is I3D-50 and takes $8\times 8$ frames as input unless specified.
        }
    \label{table:ablations}
\end{table*}

\subsection{Ablation Study}\label{subsec:ablation}
Ablation studies for the components of TPN are conducted on Kinetics-400.
Specifically, the I3D-50 backbone and the sparse sampling strategy (\ie~$8\times8$) are adopted unless specified otherwise.

\paragraph{Which feature source contributes the most to the classfication?}
As is mentioned in Sec.\ref{subsec:sources}, there exist two alternative ways to collect features from the backbone network, namely \emph{single-depth} and \emph{multi-depth}.
For the \emph{single-depth} pyramid, the output of \emph{res5} is sampled along the temporal dimension at $\left\{1,2,4,8\right\}$ intervals respectively to construct a four-level feature pyramid.
For the \emph{multi-depth} pyramid, we choose three possible combinations as shown in Table \ref{table:ablations:sources}.
The \emph{parallel flow} is adopted as the default option for feature aggregation. 
Hyper-parameters $\left\{\alpha_i\right\}_{i=1}^M$ for the \emph{multi-depth} pyramid are chosen to match its shape with the \emph{single-depth} pyramid.
For example, if \emph{res4} and \emph{res5} are selected as feature sources, the hyper-parameters will be $\left\{4, 8\right\}$.

The results of using different feature sources are included in Table \ref{table:ablations:sources},\
which suggests that the performance of TPN will drop when we take features from relatively shallow sources \eg \emph{res2} or \emph{res3}.
Intuitively there are two related factors:\
1) different from object detection where the low-level features contribute to the position regression, action recognition mainly relies on high-level semantics.
2) Another factor might be that the I3D backbone \cite{slowfast} only inflates the convolutions in the blocks of \emph{res4} and \emph{res5},
so that both \emph{res2} and \emph{res3} is unable to capture useful temporal information.
Unfortunately, inflating all 2D convolutions in the backbone will increase the computational complexity significantly and damage the performance as reported in \cite{slowfast}.
Compared to the \emph{multi-depth} pyramid, the \emph{single-depth} pyramid extracts various tempo representations by directly sampling from a single source.
Although improvement is also observed, representing video semantics only at a single spatial granularity may be insufficient.

\paragraph{How important are the information flows?}
In Sec.\ref{subsec:infoflow}, several information flows are introduced for feature aggregation.
Table \ref{table:ablations:flow} lists the performances of different information flows, keeping other components of TPN unmodified.
Surprisingly, TPN with the \emph{Isolation Flow} also boosts the performance by $0.58\%$,\
indicating that under proper modulations, the features with different temporal receptive fields indeed could help action recognition,\
even they come from a single backbone network.
TPN with the \emph{Parallel Flow} obtains the best result, leading to a performance of $76.1\%$.
The success of parallel flow suggests that lower-level features could be enhanced by higher-level features via the top-down flow for they have larger temporal receptive fields.
The semantics of higher-level features could also be enriched by lower-level features via the bottom-up flow.
More importantly, such two opposing information flows are not contradictive but complementary to each other.

\paragraph{How important are spatial semantic modulation and temporal rate modulation?}
The spatial semantic modulation and the temporal rate modulation are respectively introduced \
to overcome the semantic inconsistency in spatial dimensions and \
to adjust the relative rates of different levels in the temporal dimension.
The effect of these two modulations are studied in Table \ref{table:ablations:component}, from which we observe:
1) TPN with all the components lead to the best result.\
2) if the spatial semantic modulation contains no spatial convolutions, we have to up/down-sample the features of TPN simultaneously at spatial and temporal dimensions, \
which is ineffective for temporal feature aggregation.

\paragraph{How important is the number of input frames?}
While we use 8 frames sampled at the stride of 8 as the default input in our study experiments, we have also investigated different sample schemes.
We denote $T\times \tau$ as $T$ frames sampled with the stride of $\tau$.
And in Table \ref{table:ablations:ninput}, we include results of both I3D-50 and I3D-101 with inputs obtained by different sample schemes.
Consequently, compared to the sparser sampling scheme ($8 \times 8$), the denser sampling scheme ($32 \times 2$) tends to bring it both rich and redundant information, leading to a slight over-fitting of I3D-50.
I3D-50 + TPN, however, does not encounter such an over-fitting, obtaining an increase of $2\%$.
Moreover, consistent improvements are observed for the stronger backbone I3D-101.

\subsection{Empirical Analysis}\label{subsec:verifcation}
To verify whether TPN has captured the variance of visual tempos,
several empirical analyses are conducted on TPN.

\paragraph{Per-class performance gain \vs per-class variance of visual tempos.}
At first, we have to measure the variance of visual tempos for a set of action instances.
Unlike the concept of scale in object detection, it is non-trivial to precisely compute the visual tempo of an action instance.
Therefore, we propose a model-based measurement that utilizes the Full Width at Half Maximum (FWHM) of the frame-wise classification probability curve.
FWHM is defined by the difference between the two points of a variable where its value is equal to half of its maximum value.
We use a trained 2D TSN to collect per-frame classification probabilities for action instances in the validation set,\
and compute the FWHM for each instance as a measurement of its visual tempo,\
since when the sampling fps is fixed, a large FWHM intuitively means the action is going with a slow tempo, vice versa.
We thus could compute the variance of visual tempos for each action category.
The bottom in Figure \ref{fig:variation} shows the variances of visual tempos of all action categories,\
which reveals that not only the variance of visual tempos is large for some categories,\ 
different categories also have significantly different variances of visual tempos.

\begin{figure}[t]
    \centering
    \includegraphics[width=1\linewidth]{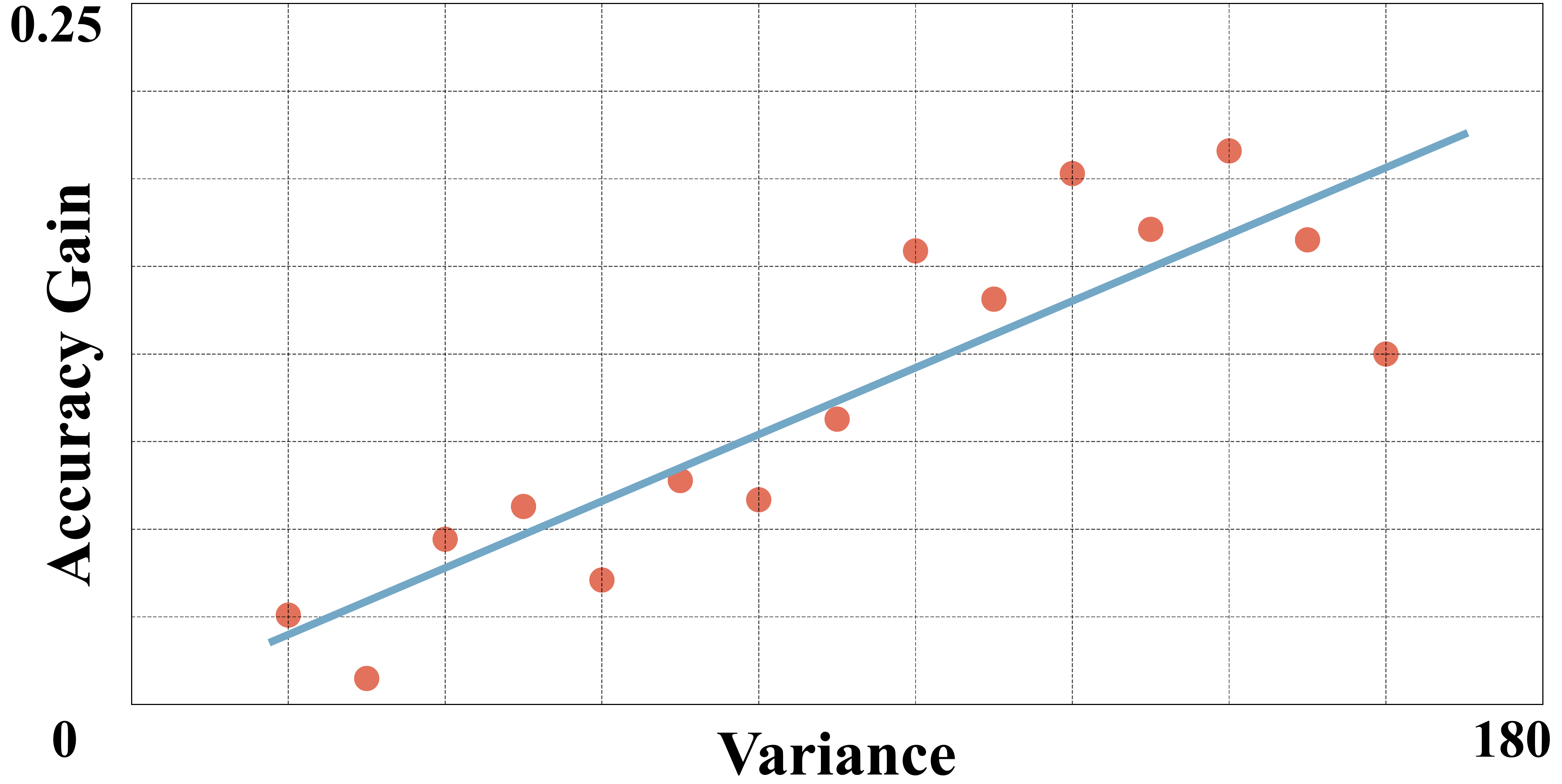}
    \caption{
        \textbf{Performance Gain \vs Variance of Visual Tempos.}
        Each \textcolor{red}{red point} denotes the mean accuracy gain within a bin of variance.
        The \textcolor{blue}{blue line} is the result of least squares approximation. 
    }
    \label{fig:correlation}
\end{figure}

Subsequently, we also estimate the correlations between per-class performance gains when adopting a TPN module and per-class variances of visual tempos.
We at first smooth the bar chart in Figure \ref{fig:variation} by dividing them into bins with an interval of 10.
We then calculate the mean of performance gains in each bin.
Finally, the statistics of all bins is shown in Figure \ref{fig:correlation},\
where performance gain is positively correlated with variance of visual tempos.
This study strongly supports our motivation that TPN could bring a significant improvement for such actions with large variances of visual tempo.

\paragraph{Robustness of TPN to visual tempo variation.}

\begin{figure}[t]
    \centering
    \includegraphics[width=1\linewidth]{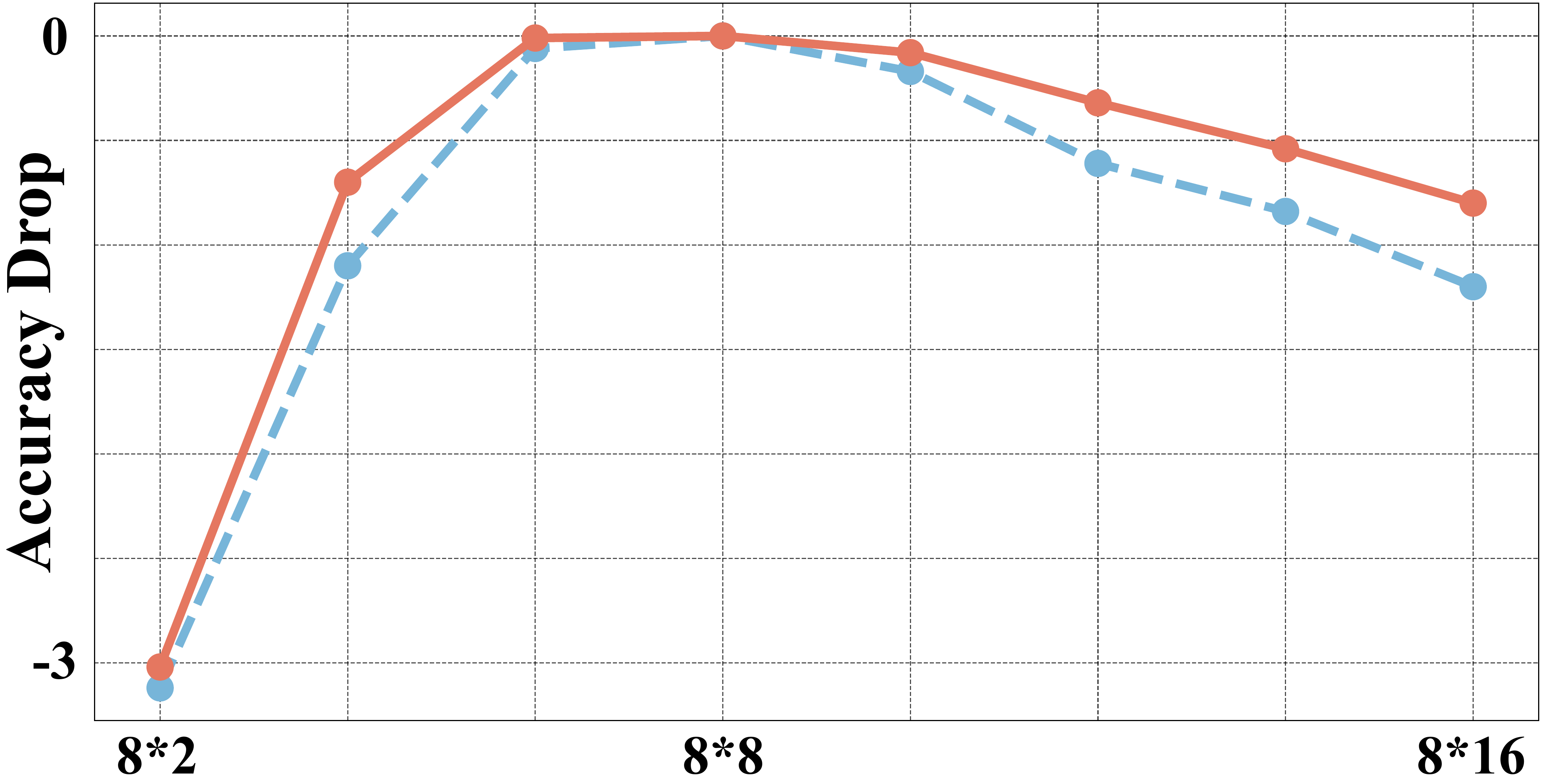}
    \caption{
        \textbf{Robustness to Variance of Visual Tempos.}
        Both baseline and TPN models are trained on $8 \times 8$ frames.
        The \textcolor{red}{red line} pictures the performance drop of baseline with TPN While the \textcolor{blue}{blue dash line} denotes that of baseline only.
    }
    \label{fig:drop_curve}
\end{figure}

Human recognizes actions easily in spite of the large variance of the visual tempos.
Does the proposed TPN module also possess such robustness?
To study this, we at first train a I3D-50 + TPN on Kinetics-400 \cite{kinetics} with $8 \times 8$ ($T \times \tau$) frames as the input.
We then re-scale the original $8\times8$ input by re-sample the frames with stride $\tau$ equals to $\{2,4,6,10,12,14,16\}$ respectively,\
so that we are adjusting the visual tempo of a given action instance.
For instance, when feeding frames sampled as $8 \times 16$ or $8 \times 2$ into the trained I3D-50 + TPN,\
we are essentially speeding up / slowing down the original action instance since the temporal scope increases/decreases relatively.
Figure \ref{fig:drop_curve} includes the accuracy curves of varying visual tempos for I3D-50 and I3D-50 + TPN,\
from which we can see TPN help improve the robustness of I3D-50, resulting in a curve with moderator fluctuations.
Moreover, the robustness to the visual tempo variation becomes clearer as we vary the visual tempo harder, as TPN could adapt itself dynamically according to the need.

\section{Conclusion}
\label{sec:conclu}
In this paper, a generic module called Temporal Pyramid Network is proposed to capture the visual tempos of action instances.
Our TPN, as a feature-level pyramid, can be applied to existing 2D/3D architectures in the plug-and-play manner, bringing consistent improvements.
Empirical analyses reveal the effectiveness of TPN, supporting our motivation and design. We will extend TPN for other video understanding tasks in the future work.

\noindent \textbf{Acknowledgments.} 
This work is supported in part by CUHK Direct Grant and SenseTime Group Limited. 
We also thank Yue Zhao for the wonderful codebase and insightful discussion.

{\small
\bibliography{references}
\bibliographystyle{plain}
}

\end{document}